# Co-speech Gesture Attribute Annotation with 3D Human Mesh Recovery

Frances Yung and Muhammad Naufilr

**Department of Language Science and Technology and Department of Computer Science**
**Saarland University, Saarbrücken, Germany.** frances@lst.uni-saarland.de muna00001@stud.uni-saarland.de

## Introduction

Co-speech gestures are an essential means of communication: performed alongside speech, hand and arm movements emphasise what is said or convey meaning that complements it [1, 2, 3]. How people produce and interpret gestures alongside speech is still largely an open question, and is an active research area of human communication. To analyze gestures, human interactions recorded in videos are categorically coded according to the semantics they represent. For example, a "sweep" from left to right meaning to "reject everything"; a sharp "chop" signalling precision, or a forward-circling hand movement to convey "progress". These gesture attributes, such as hand shape, gesture location, movement direction and speed, are all targets of analysis and have to be annotated.

Ideally human gestures are analyzed at scale, but that requires large amounts of annotated data. Annotating gestures in videos, however, is labour-intensive, typically requiring several hours of annotation for each hour of video. Manual annotation of gesture form is slow: every gesture has to be segmented and then coded for each attribute in tools such as ELAN. Several toolkits have therefore been developed to automate parts of this process. For example, SPUDNIG [4] detects when the hands move from the rest position to provide initial gesture candidates that humans can further analyze. Trujillo et al. [5] further measure the form of the movement, such as hand height, peak velocity, submovements and holds.

Existing automatic co-speech gesture analyses start from markerless pose detection, including Kinect depth recording (as in [5]) or 2D keypoint estimation from videos, such as OpenPose [6] (as in [4]), MediaPipe [7,8] or MMPose [9]. These methods locate joints in the image plane, usually with separate detectors for the body and the hands. Depth information is unavailable or only approximately estimated, so the measured geometry depends on the camera's viewpoint: a hand moving towards the camera barely moves in the image. 2D projections are limited in their ability to estimate gesture attributes that are defined relative to the speaker's body in three dimensions, such as "distance" (how far the hand is in front of the body) and "arm swivel" (how far the elbow is lifted from the side).

In this study, we instead use human mesh recovery (HMR) to estimate human pose in videos. HMR fits a 3D parametric body model to the image [10, 11]. The output is an articulated 3D skeleton that includes the hands and fingers. Its joints are positioned in 3D and follow the constraints of a human body, even when a hand is blurred or partly occluded. Specifically, we use SAM 3D Body [12], a recent HMR model that estimates body, hand and foot pose from a single frame and outputs the Momentum Human Rig [13], a representation that models the skeleton separately from body shape. SAM 3D Body's training data were collected deliberately for diversity, including unusual poses and difficult image conditions, and annotated by combining manual keypoints, multi-view geometry and model fitting [12]. This may support gesture analysis not only in laboratory recordings but also in existing videos recorded "in the wild", which are recorded at various camera angles, and where bodies move fast, overlap and are often blurred.

Figure 1 shows an example of SAM 3D Body output based on a snapshot of gesture video. Based on the upper-body shot in the video, the model trained on a large number of images can infer that the person is sitting. The full body of the gesturing person is reconstructed and can be viewed from any angle. In this study, we make use of the model's predicted skeleton to extract the relative joint positions. The complete surface mesh output is informative for more precise hand pose prediction, which we plan to explore in future work.

**Figure 1.** Example of SAM 3D Body output from a snapshot of video, taken from the GESRes dataset [14].

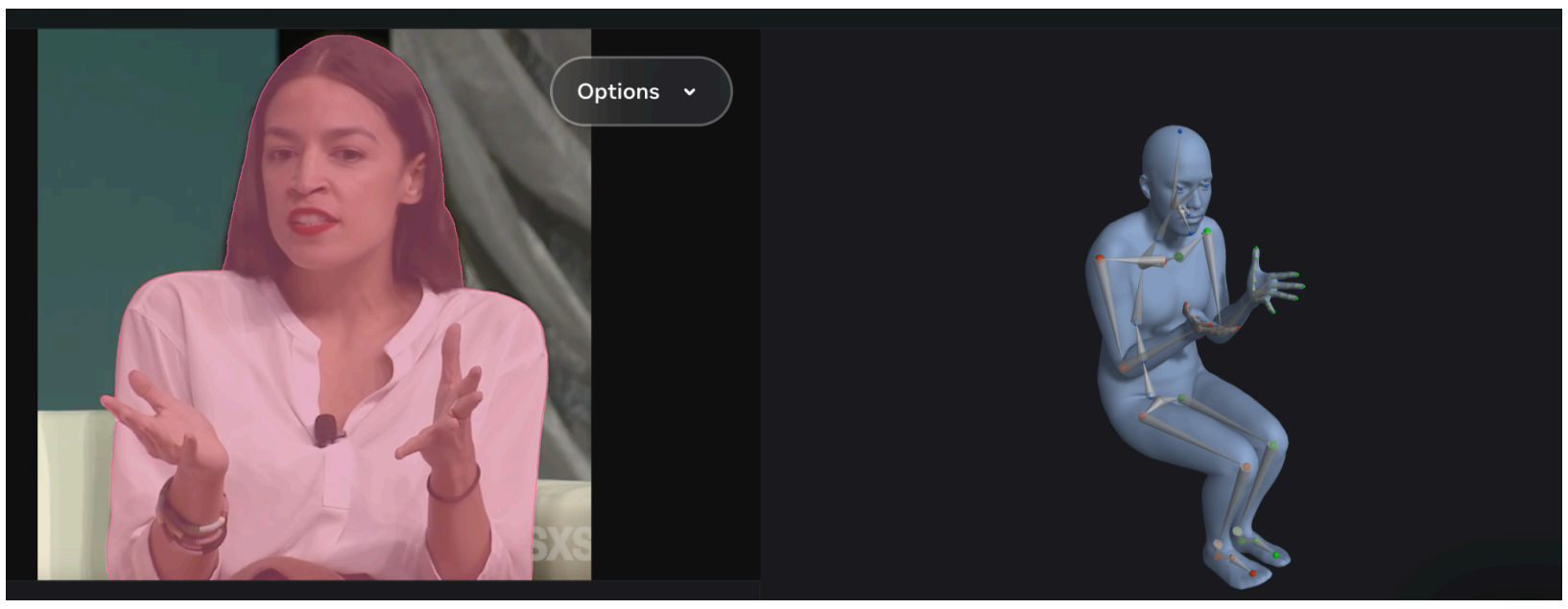

This study investigates two alternative approaches to annotating the physical attributes of co-speech gestures using HMR-based pose estimates: AI-assisted rule construction and data-driven classification using a graph convolutional network (GCN). The automated predictions are evaluated against a dataset of 2,020 manually annotated gestures, each labelled with ten physical attributes. To the best of our knowledge, this study represents the first attempt to annotate as many as ten distinct physical attributes of co-speech gestures and evaluate automated predictions against a relatively large-scale set of human annotations, reflecting a more realistic setting for practical gesture annotation.

## Methodology

### Data

We use the GESRes dataset [14], which comprises 2,373 co-speech gestures from 12 videos of nine speakers (three politicians, three clinicians, three educators). Each gesture is segmented into a separate video clip (mean duration: 2.01s, range: 0.21 - 19.26s ) and annotated manually following an annotation manual and a codebook that defines the values of each attribute. Our study focuses on predicting ten physical attributes, including 129 distinct attribute values in total. The following list presents each attribute, its number of possible values, examples, and definition.

1. **Handedness** (3 values; e.g. left, right, both): movement extent of each hand.
2. **Height** (7 values; e.g. belt, chest, shoulder, head): hand height relative to the torso.
3. **Distance** (7 values; e.g. touch, close, normal, far): arm reach and elevation; body contact.
4. **Radial orientation** (13 values; e.g. front, side, out, inward): shoulder-to-hand horizontal direction
5. **Arm swivel** (4 values; e.g. touch, normal, out, orthogonal): upper-arm angle from the side of the body.
6. **Trajectory** (4 values; e.g. straight, curved, circular, static): movement segments and turning angle.
7. **Palm orientation** (42 values; e.g. up, down, in, body, flip down to up): direction of the palm normal.
8. **Hand configuration** (42 values; e.g. palm, fist, pointing forward, steeple): finger straightness and spread, thumb contacts, etc.
9. **Movement size** (3 values; e.g. small, medium, big): extent of the hand path compared with that of other gestures produced by the same speaker
10. **Movement speed** (4 values; e.g. slow, medium, fast): peak hand speed compared with that of other gestures produced by the same speaker

In addition to these physical attributes, each gesture is also manually annotated with a *lexeme*, defined as the "overall form of the gesture" as interpreted by the annotator. Examples include "sweep" and "container". We exclude lexeme prediction because its categorization involves subjective interpretation and may depend, in part, on the accompanying speech. Instead, this study focuses on the geometric classification of the physical attributes of co-speech gestures.

**AI-assisted rule construction for gesture annotation**

The first approach classifies gesture attributes using a set of geometrical rules. Since the categories of gesture attributes, such as height and speed, are physically defined in the annotation manual, each attribute can, in principle, be classified deterministically from HMR pose estimates.

We split the gesture samples into a development set and an evaluation set. The development set (353 gestures) is used to write and tune the rules. 283 gestures were chosen with an integer program to ensure that every value of every attribute, except hand configuration, occurs at least ten times, or as often as it occurs in the full dataset if fewer than ten instances are available. We then added 70 gestures to ensure that every hand configuration value is represented at least five times. The remaining 2,020 gestures form the evaluation set. These samples are held out throughout rule development and threshold optimization and are evaluated only once, using the final rules and thresholds.

For each frame within a segmented gesture, we extract 70 joints from the SAM 3D Body output and transform their coordinates into a speaker-centred reference frame. The origin is the midpoint of the shoulders, with three axes pointing to the speaker's left, up and forward, respectively. All lengths are normalized by the speaker's shoulder width, torso length, or arm length, as appropriate. This normalization allows the rules to directly operationalize the body-relative definitions in the annotation manual (e.g., "arm swivel out = arm at 45° from the body"), independently of camera viewpoint and individual body size. Joint coordinates are smoothed using a five-frame moving average, and all frames within the annotated gesture span are considered.

Defining geometrical rules for all 129 attribute values is a complex task. We therefore adopt a human-in-the-loop approach to systematically construct and refine the rules with the assistance of a large language model (LLM) coding agent, Claude Code [15]. The process consists of three stages:

1. **Rule drafting and explanation.** The LLM agent is provided with the annotation manual, the codebook, and the dataset paper [14]. It translates each attribute value into a Python rule and generates a corresponding textual explanation. For each rule, the explanation documents the attribute definition from the codebook, the geometrical measure used at the frame level, the method for aggregating frame-level classifications into a single label for the gesture span, the procedure for combining measurements from the two hands, the numerical thresholds used to distinguish attribute values, and the sources underlying each decision.

   For example, when classifying the attribute *swivel* as *touch*, the agent generates a Python rule accompanied by the following explanation:

   - Definition: Upper arm touching the side of the body.
   - Measure: Upper-arm angle relative to the side of the body (in the frontal plane) and elbow position, measured for each frame.
   - Aggregation: The most frequently occurring class across all frames, regardless of where in the gesture span it occurs.
   - Hand combination: The most active hand only, identified by the largest bounding box of the hand-centre trajectory.
   - Sources: Swivel angles are derived from the codebook. Section 8 of the annotation manual specifies that, when a gesture position changes repeatedly, the position held for the longest duration should be selected. Selecting the most active hand is an additional design choice.

- Rule in text: The angle between the upper arm (shoulder to elbow) and the side of the body must be less than 8°, and the elbow-to-torso-side distance must be less than 0.7 times the shoulder width. Both conditions must be satisfied in the frontal view. Measurements are taken per frame and for the most active hand. The final label is assigned according to the position held for the longest duration, following the annotation manual.

The researcher reviews and comments on the generated explanations, after which the agent revises the corresponding rules. The explanatory table is regenerated after every revision to ensure that the documented rationale remains consistent with the implemented code. This process keeps the researcher responsible for the interpretation of the annotation guidelines and the validation of the resulting rules, while the LLM assists with their systematic translation into executable code.

2. **Threshold tuning.** Numeric thresholds are optimized on the development set. Only the numbers are fitted; the structure of the rules stays fixed. For each attribute, a coordinate search tries every value on a grid for one threshold at a time, keeps it if the accuracy improves, and makes two passes through all thresholds. Since the distributions of the values are skewed, with most attributes being dominated by one value (e.g. distance "normal"), the tuning process optimizes "balanced accuracy", i.e. the mean proportion of correct predictions per annotated value.

   To guard against overfitting, we perform nine-fold leave-one-speaker-out validation. In each fold, threshold optimization starts from the default values and is performed independently using the gestures of eight speakers, with predictions evaluated on the excluded speaker. The held-out predictions are pooled across all nine folds to obtain a balanced accuracy for each attribute. If the tuned thresholds improve upon the default thresholds, the attribute is subsequently tuned on the full development set of 353 gestures, and the resulting thresholds are retained.

   The threshold-tuning history is also documented in the rule explanation, e.g. *swivel_touch_degree 12 → 8; swivel_touch_distance 0.5 → 0.7.*

3. **Manual error review.** The researcher inspects the gestures in the development set that are wrongly predicted, using confusion matrices and videos with the estimated pose overlaid, and records a comment per gesture. The agent traces each error to one of three possible causes: pose estimation errors, limitations of the geometrical rules, or inconsistencies in the manual annotations.

   Based on the researcher's comments, the agent updates the rules and/or thresholds. The revisions may also override previously automatically tuned thresholds or modify rules based on the annotation manual to reflect the researcher's observations. The development set is classified again with the revised rules, and the wrong predictions are manually inspected again. This iterative process continues until the researcher has no further suggestions for revision and the explanations of the final rules are documented. The evaluation set remains untouched throughout this process, ensuring that the final evaluation provides an independent assessment of the resulting rules.

Lastly, we use the final version of the rules to predict the attributes of the 2,020 evaluation samples. The rules are not updated further. We will release the final version of the rules as an open-source tool for co-speech gesture annotation. Next, we explain the data-driven approach for gesture classification using the HMR pose estimation. We will report the performance of both approaches in the Results section.

**Data-driven gesture attribute classification with SL-GCN**

As a data-driven alternative to geometrical rule construction, we adopt the Sign Language Graph Convolutional Network (SL-GCN) architecture. SL-GCN represents a human pose sequence as a spatiotemporal graph, in which body joints form the nodes and their spatial and temporal relationships define the connections. Graph convolutional layers learn representations of the pose sequence by capturing both the spatial configuration of the body and its movement over time.

We follow the approach of Kezar et al. [16], who extend SL-GCN for sign language recognition by adding multiple classification heads to predict the phonological properties of a sign alongside its gloss. Each prediction head takes the shared pose-sequence representation as input and independently predicts one phonological feature. We adapt this architecture to the GESRes dataset by replacing the sign phonological prediction targets with the ten physical attributes of co-speech gestures, each assigned a separate classification head. The model therefore learns a shared representation of each gesture while simultaneously predicting all ten attributes in a multi-task setting.

The model is trained using the manually annotated GESRes gestures, with the classification objective defined over the ten attribute predictions. We use the same 353-gesture development set used in the rule-based approach for validation. The remaining 2,020 gestures are divided into three test folds, grouped by speaker, to ensure that each test fold contains speakers excluded from the corresponding training set. For example, Fold 1 is trained on the gestures from *Politician 2* and *3, Clinician 2* and *3, Educator 2* and *3* and tested on the gestures from *Politician 1, Clinician 1,* and *Educator 1.* The average training and testing sizes across the three folds are 1346 and 673 samples, respectively. The same validation set of 353 gestures, containing gestures from all nine speakers, is used for all folds. Each fold is trained independently for 150 epochs, and the best epoch is selected based on the mean macro F1 score across all attributes on the validation set. The best model is evaluated on its held-out test speakers. The predictions for the 2,020 test gestures are then pooled across the three folds, so that every gesture receives a prediction from a model that was not trained on that gesture or its speaker. This produces a set of predictions directly comparable with those of the rule-based approach, which also predicts the same 2,020 evaluation gestures.

To demonstrate the effectiveness of HMR for gesture annotation, we evaluate two variants of SL-GCN using different sources of pose estimates. The first uses MediaPipe pose predictions, following the input representation used in the original work of Kezar et al. (2023). The second uses SAM 3D Body pose estimates, providing the same pose information used by the rule-based approach. We compare these two variants of data-driven classification to the rule-based classification, against the expert-annotated classes in GESRes. The results are presented in the following section.

## Results

Hensel et al. [14] assessed inter-annotator agreement in GESRes using 450 gestures annotated by two annotators and Gwet's AC1/AC2 [17]. Following their approach, we compare the attribute values predicted by our methods with the corresponding GESRes annotations, using Gwet's AC1 for categorical attributes and Gwet's AC2 for ordinal attributes. The latter accounts for the degree of disagreement: for example, predicting a "normal" distance as "far" is penalized less than predicting a "close" distance as "far". Gwet's AC1 and AC2 are suitable for this dataset because traditional measures such as Cohen's kappa can over-penalize chance agreement when attribute values are highly imbalanced and dominated by a single value. Table 1 compares the performance of the three automatic classification methods. To provide a human-performance reference, we also calculate inter-annotator agreement on the 450 gestures in GESRes that were independently annotated by two annotators. For this comparison, we use the first-round annotation from each annotator, before they discussed their annotations and agreed on the final dataset labels.

**Table 1.** Gwet's AC1 (for categorical attributes) and AC2 (for ordinal attributes) between automatic predictions and the manual annotations in the GESRes evaluation set, and between two human annotators on a subset of the data. The highest automatic score for each attribute is shown in bold.

| Co-speech gesture attributes | No. of values | Type | Rules SAM 3D (N=2,020) | SL-GCN Mediapipe (N=2,020) | SL-GCN SAM 3D (N=2,020) | Human Inter-rater agreement (N=450) |
|---|---|---|---|---|---|---|
| Handedness | 3 | categorical | .600 | .595 | **.619** | .915 |
| Height | 7 | ordinal | .908 | .920 | **.936** | .963 |
| Distance | 7 | ordinal | .769 | .859 | **.906** | .873 |
| Radial orientation | 13 | categorical | .399 | .342 | **.475** | .678 |
| Arm swivel | 4 | ordinal | .908 | .920 | **.938** | .951 |
| Trajectory | 4 | categorical | .521 | **.554** | .520 | .694 |
| Palm orientation | 42 | categorical | **.320** | .223 | .239 | .627 |
| Hand configuration | 42 | categorical | **.239** | .168 | .165 | .701 |
| Movement size | 3 | ordinal | **.735** | .676 | .701 | .809 |
| Movement speed | 4 | ordinal | .861 | .935 | **.939** | .902 |
| **Mean** | | | .626 | .619 | **.644** | .811 |

Comparing the two SL-GCN models, the model using SAM 3D Body pose estimates achieves higher agreement than the model using MediaPipe pose estimates for almost every attribute. This suggests that SAM 3D Body provides more accurate pose estimates for this task. Comparing the rule-based system and SL-GCN when both use SAM 3D Body poses, each method performs better on different attributes. Overall, however, SL-GCN achieves higher mean agreement (0.644 versus 0.626). This improvement comes at the cost of requiring substantially more annotated training data: approximately five times as many gestures as the rule-based approach.

For some attributes, such as *height*, *distance*, *movement speed*, and *arm swivel*, the automatic methods achieve agreement with the manual annotations comparable to that between human annotators. For other attributes, particularly *handedness*, *radial orientation*, *trajectory*, *palm orientation*, and *hand configuration*, the agreement is lower. The relatively low performance on *palm orientation* and *hand configuration* may be partly explained by the complexity of these attributes. Human agreement is also lowest for *palm orientation*, and general-purpose body pose estimation models such as SAM 3D Body are not specialized for recognizing hand shapes, unlike models designed specifically for hand-shape recognition [12].

The low agreement for *radial orientation* is more surprising, as this attribute is geometrically defined. However, manual inspection of the samples reveals a possible explanation. *Radial orientation* includes many “A to B” values, which describe a transition between two poses over time, such as “front to side” or “inward to left”. Although the relevant positions can be defined geometrically, the annotation guidelines do not specify which time intervals within the gesture should be considered. A gesture typically contains *preparation*, *stroke*, and *retraction* phases [1]. Although these phases are not explicitly annotated in the dataset, human annotators appear to identify the stroke, which represents the central, meaningful part of the gesture, and annotate its physical attributes. The rule-based approach, by contrast, considers the entire gesture span. It may therefore identify a transition from preparation to the main stroke as an “A to B” pattern, whereas a human annotator may disregard the preparation phase and focus on the physical features of the stroke. This implicit, subjective identification of gesture phases may explain why the rule-based approach performs less well than SL-GCN on *distance* and *radial orientation*, the two attributes that include “A to B” patterns. The data-driven approach may be better able to capture these implicit temporal patterns from the training data.

In addition to these two attributes, *handedness* and *trajectory* also appear to involve subjective interpretation beyond a purely geometrical description. Specifically, their annotation may be influenced by the gesture *lexeme*. The annotation manual, for example, notes that “some lexemes by definition comprise a specific trajectory. For instance, ‘progress’ gestures are always circular.” Another example is the *lexeme* “count one hand on the other”. Although both hands visibly move in the video, the gesture is annotated as *one-handed*. Predicting these attributes may therefore require more than pose information: it may also require interpreting the gesture in relation to the meaning of the accompanying speech.

## Conclusion

In this work, we demonstrate that HMR-based body pose estimates are useful for analyzing co-speech gestures and provide more informative estimates than traditional 2D tracking methods. Within the SL-GCN framework, SAM 3D Body pose estimates generally support more accurate gesture-attribute classification than MediaPipe estimates. Given sufficient manually annotated data, data-driven classifiers can predict gesture attributes accurately when they are physically grounded and do not require subjective, multimodal interpretation.

Recent advances in LLMs make the construction and refinement of geometrical rules more accessible, offering a practical alternative to conventional data-driven approaches. In this study, the rules for 129 attribute values were developed in less than three days, using 353 manually annotated gestures for rule development and threshold optimization. Although SL-GCN achieves higher overall agreement, the rule-based approach performs better on several individual attributes. It may also be advantageous for attributes with many possible values and highly skewed distributions, as explicit rules can potentially classify rare values without requiring numerous labelled examples for each category.

Furthermore, the rule-based classification process is transparent and traceable: each decision can be linked to the annotation manual or identified as an explicit design choice. It combines LLM-assisted code generation with human oversight, allowing the researcher to evaluate conceptual validity, review classification errors, and adapt the rules to different research goals.

## Ethical statement

The current work does not involve experiments on animals or human subjects. Instead, we analyzed video data of humans from an openly available dataset, with videos that can be freely accessed. However, the pose estimation methods used in this study can reconstruct a person's body as a 3D body model without clothing, as illustrated in Figure 1. We therefore recommend obtaining informed consent from participants before applying the proposed method to analyze laboratory recordings.

## Acknowledgement

This manuscript was prepared with the help of AI software aids, including ChatGPT and Claude. Claude Code was used in the proposed method, as detailed in the Methodology section.